\documentclass[11pt]{article}
\usepackage{amsmath}
\usepackage{latexsym}
\usepackage{amsfonts}
\usepackage{bbm}

\title{Constraint Propagation\\
       as Information Maximization\thanks{
         Research Report 746, Dept. of Computer Science,
         University of Western Ontario, Canada.
       }
      }

\date{}

\author{A. Nait Abdallah\thanks{
        Department of Computer Science,
        University of Western Ontario, Canada
        and INRIA Rocquencourt, France.
        }\\
        \and M.H. van Emden\thanks{
        Department of Computer Science,
        University of Victoria, Canada.
        }
       }

\begin{document}
\maketitle

\begin{abstract}
This paper draws on diverse areas of computer science
to develop a unified view of computation:
\begin{itemize}
\item
\emph{Optimization in operations research},
where a numerical objective function
is maximized under constraints,
is generalized
from the numerical total order
to a non-numerical partial
order that can be interpreted in terms of information.
\item
\emph{Relations} are generalized
so that there are relations of which the constituent tuples
have numerical indexes, whereas in other relations
these indexes are variables.
The distinction is essential in our definition of constraint
satisfaction problems.
\item
\emph{Constraint satisfaction problems} are formulated
in terms of semantics of conjunctions
of atomic formulas of predicate logic.
\item
\emph{Approximation structures},
which are available for several important domains,
are applied to solutions of constraint satisfaction problems.
\end{itemize}
As application we treat constraint satisfaction problems over reals.
These cover a large part of numerical analysis,
most significantly nonlinear equations and inequalities.
The chaotic algorithm analyzed in the paper combines the
efficiency of floating-point computation with the correctness
guarantees of arising from our logico-mathematical model of
constraint-satisfaction problems.
\end{abstract}

\newtheorem{theorem}{Theorem}{}
\newtheorem{definition}{Definition}{}
\newtheorem{example}{Example}{}
\newtheorem{lemma}{Lemma}{}
\newtheorem{guess}{Conjecture}{}

\newcommand{\Vars}{\mathcal{X}}

\newcommand{\sq}{\ensuremath{\mbox{{\it sq}}}}
\newcommand{\sm}{\ensuremath{\mbox{{\it sum}}}}
\newcommand{\one}{\ensuremath{\mbox{{\it one}}}}
\newcommand{\lfp}{\ensuremath{\mbox{{\it lfp}}}}
\newcommand{\sbs}{\ensuremath{\sqsubseteq}}
\newcommand{\sps}{\ensuremath{\sqsupseteq}}

\def\seT#1#2{\ensuremath{ \{ #1 : #2 \}}}%
\def\set#1{\ensuremath{ \{ #1 \}}}%
\def\angTup#1{\ensuremath{ \langle #1 \rangle}}%
\def\Pwst{\ensuremath{\mathcal{P}}}

\newcommand{\restr}[2]{#1_{#2}}
\newcommand{\sqUp}[2]{{#1}^{\{#2\}}}
\newcommand{\gammaSub}[2]{\gamma_{{#1}}(#2)}
\newcommand{\cart}{\ensuremath{\mbox{\textsc{cart}}}}
\newcommand{\bx}{\ensuremath{\Box}}

\def\un{\ensuremath{\mathbf{1}}}
\def\deux{\ensuremath{\mathbf{2}}}
\def\trois{\ensuremath{\mathbf{3}}}
\def\quatre{\ensuremath{\mathbf{4}}}
\def\ord#1{\ensuremath{\mathbf{#1}}}
\def\nat{\mathbbm{N}}

\def\R{\ensuremath{\mathcal{R}}}

\def\arit#1{\ensuremath{\left| #1 \right|}}

\section{Computation as maximization in information space}

The early history of constraint processing
is written in three MIT theses:
Sutherland's, Waltz's, and Steele's \cite{sth63,wltz72,stl80}.
Already in this small selection one can discern
two radically different approaches.
Sutherland and Steele use relaxation:
starting form a guessed assignment of values to variables,
constraints are successively used
to adjust variables in such a way
as to satisfy better the constraint under consideration.
These authors followed an old idea brought into prominence
under the name of relaxation by Southwell \cite{sthwll40}.

Waltz adopted a radically different approach
(and was, to our knowledge, the first to do so).
He associated with each of the problem's variables
a \emph{domain}; that is, the set of all values 
that are not \emph{a priori} impossible.
Each constraint is then used to eliminate values from
the domains of one or more variables affected by the constraint
that are incompatible with that constraint.
In this paper we are concerned with the latter method,
which we call the \emph{domain reduction} method. 

The attraction of domain reduction
is its completeness for finite domains:
if a solution exists, then it will be found.
This in contrast with relaxation,
which can flounder forever\footnote{
But, as one may expect, domain reduction is no cure-all.
For some problems, relaxation quickly finds a solution,
and domain reduction requires an infeasible amount of time.
The $n$-queens problem for large $n$ is an example.
Van Hentenryck and Michel \cite{vhmchl05}, page 89,
mention $n = 10,000$ as a routine example for relaxation
in combination with their search technique.
}.

In this paper we present domain reduction
as an example of the view of
\emph{computation as monotonic gain of information}.
This view was pioneered by Dana Scott,
who was the first to make mathematical sense
\cite{scottPRG70} of a recursively defined function $f$.
He did this by associating with the definition of $f$
a sequence $a$ of partial functions.
If $x$ is such that $f(x)$ requires a recursion depth is at most
$n$, then $a_n(x)$ is defined and equal to $f(x)$;
otherwise $a_n(x)$ is undefined.
Thus $a$ is a sequence of partial functions
in which each function agrees with the previous one,
but is ``more defined''.

In general, if two partial functions $g$ and $h$ of the same
type are such that $h$ is defined wherever $g$ is
and such that they have the same value when both are defined,
then Scott proposed to regard $g$ as an approximation to $h$
and noted that this notion of approximation is a partial order
in the set of partial functions of the same type. 
Moreover Scott proposed to transfer the information concept
from random variables, as it was in Shannon's information theory,
to partial functions,
noting that a partial function can be regarded as containing
more information than partial functions approximating it.
The approach to the semantics of recursive definitions
can be summarized by saying that every such definition
can be regarded as the limit of a sequence of approximations
each containing more information about the limit of the sequence
than the previous one.

Scott was aware that it might seem somewhat far-fetched to give
such an interpretation to the notion of ``information''.
As a justification Scott \cite{scott72} gave another example of a
set partially ordered by information:
that of numerical intervals.
Although this certainly strengthened the case,
this suggestion has not, as far as we know, been followed up.
In this paper we do so,
motivated by the opportunities for deeper understanding
of constraint solving.

In numerical applications
the view of computation as monotonic gain of information
is more than a theoretically interesting insight:
it is adds an essential capability.
Suppose a conventional numerical computation is stopped
after 1,000 iterations and yields 1.912837465
and that it yields                1.912877134
when allowed to run for 10,000 iterations,
what do we know about the improvement obtained, if any?
If results, intermediate and final,
were expressed as intervals we would, say,
have [1.911, 1.938]\footnote{
Note the smaller number of decimals: with intervals
it becomes clear that additional decimals would be meaningless.
} after 1,000 iterations
and perhaps [1.9126, 1.9283]\footnote{
The smaller interval warrants another decimal.
} after 10,000 iterations.
Here we see that we \emph{know more} about the unknown solution
as a result of the additional computational work.
Rephrasing ``knowing more'' as ``gain in information''
suggests that the effect of iteration in interval arithmetic
can be described as ``monotonic gain of information''.
The important qualification ``monotonic'' is there
because in interval arithmetic we never need to settle for less information
as a result of additional computational work,
though we may fail to make a gain.
Moreover, such a stalling of progress
is a useful criterion for halting the iteration.

Because of the special importance of solving constraint satisfaction
problems over the reals by means of floating-point arithmetic,
we choose our example problem from this area.
Section~\ref{sec:IAvsIC} gives the needed review of interval
methods; section~\ref{sec:example} describes the example.
The new view of domain reduction as monotonic information gain
is used in Section~\ref{sec:solving}
to develop the method from first principles.
This suggests regarding the set of constraints in a constraint satisfaction
problem as a formula in predicate logic with a fixed interpretation
of predicate symbols.
The standard semantics only assigns meanings to closed formulas,
whereas here we have a formula with free variables.
Accordingly, in Section~\ref{sec:notTerm} we develop the
required extension of the semantics of predicate logic.
This needs a novel treatment of relations,
also in this section.

\section{Related work}
Following Mackworth's AC-3 algorithm \cite{mckwrth77}
there are many other papers concerned with converging
fair iterations \cite{aptEssence,bnldr97,vhmcbn98,vhsdl98,srrnpn91}.

For historical references
we refer to the textbooks \cite{dechter,aptBook2003}.

We address the connections with the work of Saraswat \emph{et al.}
\cite{srrnpn91} in Section~\ref{sec:furthWrk}.

\section{Interval arithmetic and interval constraints}
\label{sec:IAvsIC}

To facilitate the use of information in computation
we do not use interval arithmetic directly,
but indirectly via a constraint satisfaction problem
(CSP).
Such problems are solved by associating with each
unknown a \emph{set} of possible values
instead of the usual single value.
This is especially appropriate for real-valued
unknowns.
In combination with the use of floating-point arithmetic,
the sets of possible values
take the form of intervals
with floating-point numbers as bounds.
This special case of CSP solving is called
\emph{interval constraints} \cite{clr87,aptEssence}.

We introduce interval constraints by means of an example.
In interval arithmetic the rule for adding intervals is
$$
[a,b]+[c,d] = \seT{x+y}{x \in [a,b] \wedge y \in [c,d]}
$$
so that, e.g., $[0,2]+[0,2] = [0,4]$.
The analogous operation in interval constraints
starts by defining the constraint $\sm(x,y,z)$
which holds between the reals
$x$, $y$, and $z$ iff $x+y=z$.
In other words, the formula $\sm(x,y,z)$ is true
whenever $x+y=z$.
This leads to the following inference
\begin{center}
\begin{tabular}{c}
$\sm(x,y,z)$  \\
$x \in [0,2] \wedge y \in [0,2] \wedge
                      z \in [-\infty,+\infty]$  \\
\hline
$x \in [0,2] \wedge y \in [0,2] \wedge
                      z \in [0,4]$  \\
\end{tabular}
\end{center}
We use here the conventional format for inference:
the premises above the horizontal line;
the conclusion below.
The above inference coincides,
in this special case, with interval arithmetic.
Only the interval for $z$ is narrowed.

In interval constraints we may have 
\emph{a priori} constraints on \emph{all} variables,
as in
\begin{center}
\begin{tabular}{c}
$\sm(x,y,z)$  \\
$x \in [0,2] \wedge y \in [0,2] \wedge
                      z \in [3,5]$  \\
\hline
$x \in [1,2] \wedge y \in [1,2] \wedge
                      z \in [3,4]$  \\
\end{tabular}
\end{center}
Here the intervals for all three variables are narrowed.
As a result, the effect of the operation
can no longer be exclusively characterized as an addition
or as its inverse:
the effect is a mixture of several operations.
We can formulate the effect algebraically as
applying an operator,
the \emph{contraction operator}
of the constraint $\sm$,
that maps triples of intervals to triples of intervals,
in this case as
\begin{equation}\label{eq:shrinc}
([0,2],[0,2],[3,5])
\mapsto ([1,2],[1,2],[3,4]).
\end{equation}
The righthand side  of (\ref{eq:shrinc})
is the smallest triple (``box'') that can be inferred:
any box that is strictly smaller would exclude
points that are possible according the given premises of the
inference.
Thus this box is the optimal solution
to the given constraint-satisfaction problem.
The optimal solution is obtained
by one addition and two subtractions of interval arithmetic
plus a few bound comparisons.
Similarly efficient algorithms exist for some other constraints,
such as product, integer power, trigonometric and logarithmic
functions. 

We may express the contraction operator
for the \emph{sum} constraint as a mapping
from a tuple $B$ of intervals
to the least such tuple containing
the intersection of $B$ and the constraint.

In general a CSP is a conjunction of many constraints.
After applying the contraction operator for each of these once,
it is often the case that another round of applications
yields further contractions in the intervals
for some of the variables.
As the contractions are implemented
in floating-point interval arithmetic
and are assured valid by outward rounding,
there is a limit and it is reached
after a finite number of rounds of contractions.

In each of the rounds it may happen that a constraint is found
that does not contain variables for which a bound has changed.
In such cases the contraction operator
for that constraint has no effect and can be skipped.
Algorithms have been developed
that perform such optimizations \cite{aptEssence}.

\section{An example of solving by interval constraints}
\label{sec:example}

Let us consider the problem of determining the intersection
points of a parabola and a circle.
For example, to solve the system
\begin{equation}\label{eq:original}
  \begin{array}{rcl}
     y         &=& x^2 \\
     x^2 + y^2 &=& 1  
  \end{array}
\end{equation}
with $x \in [0,1]$ and $y \in [0,1]$.
One can eliminate $y$ and solve instead
$x^4 + x^2 = 1$,
which has two real roots.
However, for the purpose of illustrating
solving by interval constraints,
we ignore this opportunity for simplification
and we numerically solve the original system
(\ref{eq:original}).

The method of interval constraints applies to a class
of constraints in the form of
equalities or inequalities between real-valued expressions.
The \sm\ constraint in Section~\ref{sec:IAvsIC} is an example:
it takes the form of the equation $x+y=z$.
As we mentioned in that section, there is an efficient
implementation of the optimal contraction operator for it.

The second equation in (\ref{eq:original}) is not primitive;
it has to be transformed
to an equivalent set of primitive constraints.
In this example the primitive constraints
\sq, \sm, and \one\ are needed.
The constraint $\sq(u,v)$ is defined as $u^2 = v$,
$\sm(u,v,w)$ is defined as $u+v=w$,
and $one(u)$ is defined as $u=1$.
In this way (\ref{eq:original}) becomes the
following set of constraints:
\begin{equation}\label{eq:constraints}
\{\sq(x,y), \sq(y,z), \sm(y,z,u), \one(u)\}.
\end{equation}
The unknowns $x$, $y$, $z$, and $u$ are real numbers.
The introduction of $z$ and $u$ is the result of reducing
the given constraints to primitive ones.
In more typical cases the given constraints are so complex
that the introduced variables greatly outnumber the original ones.

In the example it is given that
$x$, $y$ and $z$
satisfy the above constraints.
From the original problem statement we have in addition
that $x \in [0,1]$ and $y \in [0,1]$.
Of the auxiliary unknown $z$
we initially know nothing:
$z \in [-\infty,+\infty]$ and
$u \in [-\infty,+\infty]$.

In effect, we have transformed
(\ref{eq:original})
to the system
\begin{eqnarray}\label{eq:deriv}
  \begin{array}{rcl}
     y         &=& x^2  \\
     z         &=& y^2  \\
     y + z     &=& u    \\
     u         &=& 1  
  \end{array}
\end{eqnarray}

Instead of solving the original system 
(\ref{eq:original})
we solve equivalently the constraints
(\ref{eq:constraints}).
This is done by repeatedly applying in arbitrary order
the contraction operators until
there is no change in any of the intervals
associated with the unknowns.
Applying the contraction operators of $\sq(y,z)$ and $one(u)$ results
in a drastic narrowing of the intervals for $z$ and $u$:
they change from 
$[-\infty,+\infty]$ to $[0,1]$ for $z$ and to $[1,1]$ for $u$.
After this, none of the contraction operators
of the four constraints results in a change.
Therefore this is as far
as contraction operator application can take us.

To obtain more information about possibly existing solutions,
we split the CSP with interval $X = [0,1]$ for unknown $x$
into two CSPs that are identical
except for the intervals of $x$.
In the first CSP the interval for $x$ is the left half of $X$;
in the second CSP it is the right half.
Then we start another round of contraction operator applications
starting from one of the halves as initial box:
\begin{equation} \label{eq:box1}
x \in [0,\frac{1}{2}],
y \in [0,1],
z \in [0,1],
u \in [1,1].
\end{equation}
Applying the contraction operator
for $\sq(x,y)$ results in $y \in [0,1/4]$.
Applying the contraction operator
for $\sq(y,z)$ results in $z \in [0,1/16]$.
Applying the contraction operator
for $\sm(y,z,u)$ results in $u \in [0,5/16]$.
Applying the contraction operator
for $one(u)$ causes the interval for $u$ to become empty.
This proves that there is no solution in the initial box
(\ref{eq:box1}).

We now turn to the other half:
\begin{equation} \label{eq:box2}
x \in [\frac{1}{2},1],
y \in [0,1],
z \in [0,1],
u \in [1,1].
\end{equation}
Applying the contraction operator for $\sq(x,y)$ results in
$y \in [\frac{1}{4}, 1]$.
Continuing in tabular form gives
\begin{center}
\begin{tabular}{c|c|c|c|c}
        & \multicolumn{4}{c}{Interval} \\ \cline{2-5}
        & $x$ & $y$ & $z$ & $u$ \\ \cline{2-5}
        & $[0.5,1]$ & $[0,1]$ & $[0,1]$ & $[1,1]$ \\
\hline
Apply   &           &         &         &                     \\
\cline{1-1}
$\sq(x,y)$   &   &   [$\frac{1}{4}$,1]    &  & \\        
$\one(u)$   &   &                         &  & $[1,1]$ \\        
$\sm(y,z,u)$ &   &       &   [0,$\frac{3}{4}$]  &   \\        
$\sq(y,z)$   &   & $[\frac{1}{4},\frac{1}{2}\surd 3]$
             & $[\frac{1}{16},\frac{3}{4}]$ &     \\        
\end{tabular}
\end{center}

Now the intervals for $x$ and $y$ continue getting smaller
until the least floating-point box has been reached
that contains a solution:
the intervals for $x$ converge to
a small interval containing
$\surd (\frac{1}{2}(\surd 5 - 1))$,
while the intervals for $y$ converge to
a small interval containing
$\frac{1}{2} (\surd 5 -1 )$.

\section{Notation and terminology for relations and constraints}
\label{sec:notTerm}

We take it that (\ref{eq:constraints})
is intuitively clear,
but how do we characterize mathematically
any solutions that such a CSP may have
and how do we characterize mathematically
an algorithm for obtaining such a solution?
Consider for example the constraints $sq(x,y)$ and $sq(y,z)$.
They clearly have something in common: $sq$,
which must be some kind of relation.
But the constraints are different from each other
(otherwise their conjunction could be simplified
by dropping either of them)
and also different from $sq$, whatever \emph{that} may be.

In this section we develop
a set-theoretic formulation of constraint-satisfaction problems
and illustrate it by the example in Section~\ref{sec:example}.
We find that such a formulation is facilitated
by a treatment of relations and operations on them
that is in the spirit of the conventional treatment,
but differs in details.
In particular, we need to clarify the difference
between relations and constraints
as well as the connection between these.

\subsection{Functions}
We denote by $S \to T$ the set of total functions
that are defined on $S$ and have values in $T$.
If $f \in (S \to T)$ we say that $f$
``has type'' $S \to T$.
If $S' \subseteq S$, then we define $f_{S'}$,
the \emph{restriction} of $f$ to $S'$
as the function in $S' \to T$
such that for all $x \in S'$
we have $f_{S'}(x) = f(x)$.

\subsection{Tuples}
As is the case conventionally,
our relations are sets of tuples of the same arity.
However, we need the possibility to index tuples
either by variables
or by the conventional indexes \set{0,1,2, \ldots}.
Hence we define a tuple
as an element of the function set $I \to T$,
where $I$ is an arbitrary set to serve as index set.
$I \to T$ is the \emph{type} of the tuple.

\emph{Example}
If $t$ is a tuple in $\set{x,y} \to \R$,
then we may have $t(x) = 1.1$ and $t(y) = 1.21$. 

\emph{Example}
$t \in \trois \to \set{a,b,c}$,
where $\trois = \set{0,1,2}$
and $t(0) = b$, $t(1) = c$, and $t(2) = c$.
In cases like this, where the index set is an ordinal,
we use the compact notation $t = [b,c,c]$.
In general, we write \ord{n}\ for $\set{0,\ldots, n-1}$.

When a function is regarded as a tuple,
then the restriction operation on functions
is called \emph{projection}.
E.g. if $t = [2,1,3]$ and $t' = t_{\{0,2\} }$,
then $t'(0) = 2$ and $t'(2) = 3$;
$t'(1)$ is not defined.

\subsection{Approximation structures}

In \cite{scott72} Dana Scott proposed that computation steps
be viewed as transitions in a partially ordered space of data.
In his view computation consists of generating a time-ordered
sequence $d_0, d_1, d_2, \ldots$ with the property that
the successive data $d_i$ are each approximated by the previous
in the sense of holding information
about the limit of the sequence
that is compatible and is at least as informative.
We write
$d_0 \sqsubseteq d_1 \sqsubseteq d_2 \sqsubseteq \cdots$ 
where $\sqsubseteq$ is the partial order.

Scott was primarily interested in using his approach
to model mathematically the evaluation of recursively
defined functions. This requires mathematically rather
sophisticated constructions.
However, the idea also applies to situations
covered by the following definition.

\begin{definition}\label{def:apprStruct}
An \emph{approximation structure} for a set $D$
is a set $A$ of subsets of $D$ such that 
(1) $A$ is closed under finite intersection, 
(2) $A$ is closed under intersection of (possibly infinite)
$\subseteq$-descending chains of subsets,
and (3) $A$ contains $D$ as an element.
The information order $\sqsubseteq$ of $A$ is defined
as the inverse of the inclusion $\subseteq$ of subsets.

An \emph{approximation domain} is a pair $\langle D, A \rangle$
formed by a set $D$ and an approximation structure $A$ on $D$.
It turns out to be tiresome to say
``an approximation domain $(D,A)$ for some $A$'',
so that we may speak of
``an approximation domain $D$''
when no ambiguity arises regarding $A$.
\end{definition}

\begin{lemma}\label{lem:leastElt}
If $D' \subseteq D$,
then there exists in any approximation structure for $D$
a $\subseteq$-least element containing $D'$.
\end{lemma}

\begin{definition}
If $A$ is an approximation structure for $D$,
then for $D' \subseteq D$ we define $\alpha_A(D')$
to be the least element of $A$ containing $D'$.
\end{definition}
The set $\alpha_A(D')$
corresponds to the maximum amount of information about $D'$ that is
expressible within approximation structure $A$.

\emph{Example}
The intervals form an approximation structure
in the set $\R$ of real numbers,
where we define an interval as
$\set{x \in \R: a \leq x \leq b}
$,
where
   $a \in \R \cup \set{-\infty}$
and
   $b \in \R \cup \set{+\infty}$.
We write $[a,b]$ for this interval.
Note that with this definition,
e.g., $+\infty \not \in [0,+\infty]$.

\emph{Example}
Let $F$ be a subset of the set $\R$ of reals.
The $F$-intervals are an approximation structure in $\R$,
where an $F$-interval is 
$\set{x \in \R : a \leq x \leq b}$
where
   $a \in F \cup \set{-\infty}$
and
   $b \in F \cup \set{+\infty}$.
An important case: $F$
is the set of finite double-length IEEE-standard
floating-point numbers.
The latter include $-\infty$ and $+\infty$,
so that pairs of these numbers
are a convenient representation
for the elements of this approximation structure. 

\subsection{Relations}
A relation is a set of tuples with the same type.
This type is the \emph{type} of the relation.

If $r$ is a relation with type $I \to T$,
then the \emph{projection} of $r$ on $I' \subseteq I$
is $\set{f' \in I' \to T :
  \exists f \in r. f_{I'} = f' 
}$
and denoted $\pi_{I'}r$.

\emph{Example}\\
$\sm = \set{[x,y,z] \in (\trois \to \R) : x+y=z}$
is a relation of type $\trois \to \R$,
where $\trois = \{0,1,2\}$.
Compare this relation to the relation
$\sigma = \set{s \in (\set{x,y,z} \to \R) : s_x+s_y=s_z}$.
As their types are different,
they are different relations;
$[2,2,4] \in \sm$ is not the same tuple
as $s \in \sigma$ where
$s_x=2$,
$s_y=2$, and
$s_z=4$.

\emph{Example}\\
If $S$ has one element,
then a relation of type $S \to T$ is a \emph{unary} relation.
Such a relation is often identified with a subset of $T$.
For example, for $a$ in $\R \cup \set{-\infty}$
and $b$ in $\R \cup \set{+\infty}$,
$\set{f \in (\set{x} \to \R) : a \leq f_x \leq b}$
is a unary relation that
is often identified with the interval $[a,b]$.
Maintaining the distinction between the two is important in the
current setting (see Section \ref{sec:constraints}).

\begin{definition}
If $r_0$ and $r_1$ are relations
with types $I_0 \to T$ and $I_1 \to T$, respectively,
then the \emph{join} $r_0 \Join r_1$ of
$r_0$ and $r_1$ is
$$\set{f \in (I_0 \cup I_1) \to T :
 f_{I_0} \in r_0 \mbox{ and }
 f_{I_1} \in r_1
}.$$

The join of relations that have disjoint index sets
is called the \emph{product} of these relations.
\end{definition}

We avoid the term ``Cartesian product'' because 
it is usually understood to consist of tuples with index
set $\{0,\ldots,n-1\}$ for some natural number $n$.

\begin{definition}
Let $r$ be a relation of type $I \to T$ and let $I \subseteq J$.
Then we write the \emph{cylinder} on $r$ with respect to $J$
as $\pi_J^{-1} r$ and define it as
the greatest relation $g \subseteq (J \to T)$
such that $\pi_I g = r$.
\end{definition}

Cylindrification is inverse to projection in the sense that
$\pi_I(\pi_J^{-1} r) = r$.

\begin{definition}
Let $I = \{i_0, \ldots, i_{n-1}\}$ be an index set.
A \emph{box} is a product of unary relations
$
r_0 \subseteq \{i_0\} \to D
,\ldots,
r_{n-1} \subseteq \{i_{n-1}\} \to D
$.
In case $ r_0 ,\ldots, r_{n-1} $ are intervals,
then one may refer to the box
as an \emph{interval box}.
\end{definition}


%
%

\subsection{Boxes as approximation domain}\label{sec:boxApprox}

\begin{lemma}
Let $I = \set{i_0,\ldots,i_{n-1}}$ be a finite index set
and let $B$ be the set of boxes of type $I \to D$.
Then $\angTup{I \to D, B}$ is an approximation domain.  
\end{lemma}
\emph{Proof}
We need to show the three defining properties
(Definition~\ref{def:apprStruct}).
In this case one can show closure under arbitrary
finite or infinite intersection,
so that the first two properties can be established simultaneously.

Let \seT{r^j}{j \in J} be a possibly infinite family of boxes,
$r^j = r_0^j \Join \cdots \Join r_{n-1}^j$,
with $r^j_k \subseteq \set{i_k} \to D$ for all $k \in \ord{n}$.

Let
$$ r
= \bigcap_{j \in J} r^j
= \bigcap_{j \in J} (r_0^j \Join \cdots \Join r_{n-1}^j).
$$
Then
\begin{eqnarray*}
f \in r = \bigcap_{j \in J} r^j 
   & \Leftrightarrow & \forall j \in J. \; f \in r^j 
   \;\Leftrightarrow\; \forall j \in J. \; \forall k \in \ord{n}. \;
     f_{i_k} \in r_k^j \\
& \Leftrightarrow & \forall k \in \ord{n}. \; \forall j \in J. \;
     f_{i_k} \in r_k^j
     \;\Leftrightarrow\;  \forall k \in \ord{n}. \;
     f_{i_k} \in \bigcap_{j \in J} r_k^j \\
& \Leftrightarrow &
     f \in \bigcap_{j \in J} r_0^j \Join \cdots \Join
               \bigcap_{j \in J} r_{n-1}^j\\
\end{eqnarray*}
Hence
$$
\bigcap_{j \in J} r^j
= \bigcap_{j \in J} r_0^j \Join \cdots \Join \bigcap_{j \in J} r_{n-1}^j
$$
is also a box,
so that the intersection of a possibly infinite family of boxes
is a box.

We finally need to show that the full relation $r = I \to D$ is a box.
Letting $r_k = \set{i_k} \to D$,
we have that
\begin{eqnarray*}
I \to D &=& \\
\set{i_0, \ldots, i_{n-1}} \to D & = & \\
(\set{i_0} \to D) \Join \cdots \Join (\set{i_{n-1}} \to D) &=& \\
r_0 \Join \cdots \Join r_{n-1}
\end{eqnarray*}
is a box.
\hfill $\Box$

\paragraph{}

Therefore, for every relation $r$
of type $\set{i_0,\ldots,i_{n-1}} \to D$
there is a least box containing $r$,
which justifies the following definition.

\begin{definition}
The \emph{box operator} applied to a relation
$r$ with type $\set{i_0,\ldots,i_{n-1}} \to D$
is the least box $\bx r$ that contains $r$.
\end{definition}

\subsection{Constraints}\label{sec:constraints}
A constraint is a syntactic entity
that is used to denote a relation.
A constraint has the form of an atomic formula
in a theory of predicate logic without
function symbols.
The semantics of predicate logic assigns a relation $r$
to an atomic formula
$p(q_0, \ldots, q_{n-1})$ with set $V$ of variables.
The relation $r$ depends on the interpretation of $p$
and on the tuple $[q_0, \ldots, q_{n-1}]$ of arguments.
These arguments are variables,
not necessarily all different.
The first-order predicate logic interpretation of the
language of atomic formulas,
which identifies the argument occurrences by numerical indexes,
forces $\ord{n} = \set{0,\ldots, n-1}$
to be the index set of the relation $M(p)$,
the relation that is the meaning of the predicate symbol $p$
under the given interpretation.
In our setting, instead, the index set associated with 
the constraint denoted by 
$p(q_0, \ldots, q_{n-1})$ is the set $V$ of 
distinct variables occurring in atomic formula
$p(q_0, \ldots, q_{n-1})$.

The interpretation $M$ that assigns a relation
of type $\set{0,\ldots, n-1}$ to an $n$-ary predicate symbol $p$
needs to be extended to an interpretation $M$
that also assigns a relation of type $V \to D$
to a constraint.

\begin{definition}
Let $c = p(q_0, \ldots, q_{n-1})$
where $V$ is the set of variables in
$\{q_0, \ldots, q_{n-1}\}$.
We define
$$ M(c) =
\set{a \in V \to D : [a(q_0), \ldots, a(q_{n-1})] \in M(p)}.
$$
\end{definition}

As a result of this definition the meaning of a constraint $c$
with set $V$ of variables is a relation of type $V \to D$.
One can view the argument tuple of a constraint as an
operator that converts a relation $M(p)$ of type $\ord{n} \to D$
to relation $M(c)$ of type $V \to D$.
This is an extension of the usual semantics of predicate logic.

\paragraph{}
\emph{Example}\\
Let $\sq$ be the binary relation over the reals
where the second argument is the square of the first.
That is,
$M(sq) = \set{f \in (\set{0,1} \to \R): f_1 = f_0^2}$.
The constraints
$\sq(x,y)$,
$\sq(y,x)$, and
$\sq(x,x)$
denote different relations,
as we verify below.

Given that
$M(sq) = \set{f \in (\set{0,1} \to \R): f_1 = f_0^2}$
we have
\begin{eqnarray*}
M(sq(x,y)) &=&
\set{a \in (\{x,y\} \to \R) : [a(x),a(y)] \in M(sq)} \\
           &=&
\set{a \in (\{x,y\} \to \R) : a(x)^2 = a(y)} \\
M(sq(y,x)) &=&
\set{a \in (\{x,y\} \to \R) : [a(y),a(x)] \in M(sq)} \\
           &=&
\set{a \in (\{x,y\} \to \R) : a(y)^2 = a(x)}  \\
M(sq(x,x)) &=&
\set{a \in (\{x\} \to \R) : [a(x),a(x)] \in M(sq)} \\
           &=&
\set{a \in (\set{x} \to \R) : a(x) = 0 \vee a(x) = 1}
\end{eqnarray*}

\begin{definition}
  A tuple $ f \in V \to D $ satisfies a constraint $c$ if
  and only if the restriction of $f$ to the set of variables occurring
  in $c$ belongs to $M(c)$.
\end{definition}

\subsection{Constraint-satisfaction problems}

\begin{definition}
A \emph{constraint-satisfaction problem} (CSP)
has the form $\angTup{C,V,D,M}$
and consists of
a set $C = \set{s_0,\ldots,s_{m-1}}$ of constraints,
a set $V$, which is the set of the variables
occurring in the constraints,
a set $D$, the \emph{domain} of the CSP, and
an interpretation $M$,
which maps every $n$-ary predicate symbol occurring in
any of the constraints to a relation of type $\ord{n} \to D$.
A \emph{solution} to $\angTup{C,V,D,M}$
is $a \in V \to D$ such that
$a_{V_i} \in M(s_i)$ for all $i \in \ord{m}$,
where $V_i$ is the set of variables in $s_i$.
\end{definition}

It follows that the set $\sigma$ of solutions of the CSP
is a relation of type $V \to D$.

\emph{Example}
In $\angTup{C,V,D,M}$,
let $C = \{\sq(x,y), \sq(y,z), \sm(y,z,u), \one(u)\}$
(Equation (\ref{eq:constraints})),
$V = \set{x,y,z,u}$,
$D = \R$,
$M(\sq) = \{f \in (\{0,1\} \to \R) : f(1) = f(0)^2\}$,
$M(\sm) = \{f \in (\{0,1,2\} \to \R) : f(2) = f(0)+f(1)\}$,
and $M(\one) = \{f \in (\{0\} \to \R) : f(0) = 1\}$.
The set $\sigma$ of solutions
is a relation $\sigma \subseteq V \to \R$
such that
$\pi_{\set{x,y}} \sigma = \set{p_0,p_1}$
where
$p_0(x) = - \surd (\frac{1}{2}(\surd 5 - 1)) $,
$p_0(y) = \frac{1}{2} (\surd 5 -1 )$,
$p_1(x) =  \surd (\frac{1}{2}(\surd 5 + 1)) $,
and
$p_1(y) = \frac{1}{2} (\surd 5 -1 )$.

This example shows a CSP with a finite and small solution set.
Sudoku puzzles are another such example.
It often happens that the solution set has an infinite number of
elements, or a finite number that is too large to list
or to process on a computer.

\begin{theorem}\label{thm:joinSigmas}
Let $\sigma$ be the solution set of a CSP 
$C = \set{s_0,\ldots,s_{m-1}}$ with $M$ as interpretation
for its predicate symbols.
Then we have
$$
\sigma = M(s_0) \Join \cdots \Join M(s_{m-1}).
$$
\end{theorem}
\emph{Proof} 
By induction on the size of set
$\set{s_0,\ldots,s_{m-1}}$.
The base case $ C = \set{s_0}$ 
is trivial.

Assume that the theorem holds for a constraint set
$C_k = \set{s_0,\ldots,s_{k-1}}$
of size $k \geq 1$, and let 
$\sigma(C_k) = M(s_0) \Join \cdots \Join M(s_{k-1})$
denote the solution set of $C_k$. Consider constraint set
$C_{k+1} = C_k \cup \set{s_k}$.
Any tuple $t$ which is a solution of $C_{k+1}=  C_k \cup \set{s_k}$
must be such that the restriction of $t$ to the set of 
variables occurring in $C_k$ is a solution of $C_k$, and
the restriction of $t$ to the set of variables occurring in 
$s_k$ is a solution of $s_k$.
Whence $\sigma(C_{k+1}) \subseteq \sigma(C_k) \Join M(s_k)$.
Conversely, if $ t \in \sigma(C_k) \Join M(s_k)$, then by
construction $t$ satisfies $C_k$ as well as $s_k$, whence $t$
satisfies $C_{k+1} = C_k \cup \set{s_k}$.
Therefore $\sigma(C_{k+1}) = \sigma(C_k) \Join M(s_k)$.
\hspace{\fill}$\Box$

\section{Solving constraint-satisfaction problems}
\label{sec:solving}

What does it mean to ``solve'' a CSP?
It is rare for the solution set $\sigma$
to have but few elements, as it does in Sudoku.
Though occupying only a small proportion of the type,
$\sigma$ may have a finite
and overwhelmingly large number of elements;
it may also be an infinite set.
Hence we can typically only hope to obtain
\emph{some} information about $\sigma$. 
Useful information can come in the form of an \emph{approximation}.

If the approximation domain consists of computer-representable
sets, as it typically does,
then $\Box \sigma$ is computer-representable,
but will usually give too little information about $\sigma$.
But $\Box \sigma$ is useful in case one can show
that it is empty:
in that case $\sigma$ is empty;
i.e. the CSP has no solutions.
This is an advantage of treating
numerical problems as CSPs:
in conventional computation one can only conclude that no
solutions were found.
By formulating the problem
as a CSP with intervals as approximation structure
one may be able to prove that no solutions exist.
The possibility of proof of non-existence by means of standard
floating-point arithmetic (and all its rounding errors)
is a valuable complement to conventional numerical analysis.

In case it is not possible
to show that $\Box \sigma$ is empty,
one subdivides the box under consideration
and one may be able to show
that one of these subdivisions has no solutions.
Let box $P$ (``probe'') be such a subdivision.
We use it to reduce the partial solution of
the problem of determining $\sigma$
to that of determining any solutions that might occur in $P$,
or to find, also usefully, that no solutions occur in $P$.
Thus we proceed to obtain information about $\sigma \cap P$.
This intersection is in general not a box,
so is not necessarily computer-representable.
Hence it is an appropriate task for an algorithm
to determine $\Box (\sigma \cap P)$
for a given CSP and a suitable $P$,
or an approximation to $\Box (\sigma \cap P)$
(which is itself an approximation).

Subdivision of $P$ should result in subsets of $P$
whose union includes $P$.
These subsets are subject to the same consideration:
if absence of solutions cannot be shown
and if amenable to subdivision,
the process repeats for such a subset.
Any box $P$ defines a tree of subsets
to be processed in this way:
solving a CSP requires,
in addition to an attempt to show
the absence of solutions in a given box,
a search over the tree of subboxes of the initially given box.
The ``solution'' of a numerical CSP
is necessarily a list of boxes
each of which is too small to subdivide
and of which the absence of solutions cannot be shown.
Of a solution $x \in \R^n$
the best one can typically do is to fail to show
that $\Box(\{x\})$ contains no solutions of the CSP.

\subsection{Contraction operators}
A contraction operator transforms a box $B$
into a box $B' \subseteq B$
such that there is no solution in $B \setminus B'$.
Two kinds of contraction operators on boxes are defined here:
operators defined by relations, and operators defined by constraints.

\subsubsection{Contraction operators defined by a relation}
\label{sec:Contraction_relation}
\begin{definition}\label{def:gamma}
Let $D$ be an approximation domain and $I$ an index set.
Any relation $r$ of type $I \to D$
determines the mapping $\gamma_r(P) = \Box(r \cap P)$,
the \emph{contraction operator} of $r$,
that maps boxes with type $I \to D$
to boxes with the same type.
\end{definition}

Benhamou and Older \cite{bnldr97} introduced this formula
for intervals of reals.
Here it is generalized to approximation systems in general.

\begin{lemma}\label{lem:gamma:properties}
The contraction operator $\gamma_r$
is idempotent, monotonic, inflationary and correct.
\end{lemma}
\emph{Proof}
We have that
$
\Box(\Box(r \cap P) \cap P)
=
\Box(r \cap P) \cap P
=
\Box(r \cap P)
$;
hence $\gamma_r$ is idempotent.

$\Box$ is monotonic and intersection is monotonic in both
arguments, so $\gamma_r$ is monotonic.

$\gamma_r(P) = \Box(r \cap P) \subseteq \Box P = P$,
so that
$P \sbs \gamma_r(P)$.
That is, $\gamma_r$ moves up in the
(information) partial order: $\gamma_r$ is inflationary.

We have that
$r \cap (P \setminus \gamma_r(P))) = \emptyset$
meaning that $\gamma_r$ is correct in the sense that
it does not remove any part of $r$ from its argument.

\paragraph{An example of a contraction operator}

The contraction operator for the $\sm$ constraint
acting on a box
$$
(\set{x} \to [a,b]
  )\Join( \set{y} \to [c,d]
  )\Join( \set{z} \to [e,f])
$$
where
$a,b,c,d,e,f$ are finite
IEEE-standard floating-point numbers is given by
\begin{multline*}
\gamma_{M(sum(x,y,z))}
((\set{x} \to [a,b]
  )\Join( \set{y} \to [c,d]
  )\Join( \set{z} \to [e,f]))  = \\
(\set{x} \to [a',b']
  )\Join( \set{y} \to [c',d']
  )\Join( \set{z} \to [e',f']).
\end{multline*}
Here 
\begin{eqnarray*}
\; [a',b'] &=& [a,b] \cap [(e-d)^-,(f-c)^+] \\
\; [c',d'] &=& [c,d] \cap [(e-b)^-,(f-a)^+] \\
\; [e',f'] &=& [e,f] \cap [(a+c)^-,(b+d)^+] \\
\end{eqnarray*}
where
superscript $^-$ means that the floating-point
operation is performed in round-toward-minus-infinity
mode
and
superscript $^+$ means that the floating-point
operation is performed in round-toward-plus-infinity
mode.
In this way correctness of $\gamma_{sum}$
is maintained in the presence of rounding errors.

In Equation~(\ref{eq:shrinc}) the contraction operator
is applied in the case where
$a = 0$,
$b = 2$,
$c = 0$,
$d = 2$,
$e = 3$,
and
$f = 5$.
Applying
$\gamma_{M(sum(x,y,z))}$
in this special case gives
\begin{center}
\begin{tabular}{lll}
$[a',b']$ & $=\;\; [0,2] \cap [1,5]$ & $=\;\; [1,2]$\\
$[c',d']$ & $=\;\; [0,2] \cap [1,5]$ & $=\;\; [1,2]$\\
$[e',f']$ & $=\;\; [3,5] \cap [0,4]$ & $=\;\; [3,4]$\\
\end{tabular}
\end{center}

This only gives the general idea.
A practical algorithm has to take care of the possibility
of overflow.
It also has to allow for the possibility
that $a,c$ or $e$ are $-\infty$
and
that $b,d$ or $f$ may be $+\infty$
so that the undefined cases
$(+\infty) - (+\infty)$,
$(-\infty) + (+\infty)$,
and
$(+\infty) + (-\infty)$
have to be circumvented.
For details about such algorithms see \cite{hckvnmdn01}.

\subsubsection{Contraction operators defined by a CSP}
\label{sec:Contraction_CSP}
In the CSP defined by the constraints $\set{s_0,\ldots,s_{m-1}}$,
let us write $\sigma_i$ for $M(s_i)$.
Then Theorem~\ref{thm:joinSigmas} says that
$$
\sigma = \sigma_0 \Join \cdots \Join \sigma_{m-1}.
$$
The $\gamma$ operator of Definition~\ref{def:gamma}
is not useful for $r = \sigma$,
but it can be useful for the $r = \sigma_i$,
the solution sets for the constraints by themselves.
In fact, the constraints are chosen to be such
that one has an efficient algorithm for each $\gamma_{\sigma_i}$.

\begin{definition}\label{def:bigGamma}
Let
$\angTup{\set{s_0,\ldots,s_{m-1}},V,D,M}$
be a CSP.
Let $\sigma_i = M(s_i)$ and let
$V_i$
be the set of variables of $s_i$.
We define
$$\gamma_i(P) = \pi_V^{-1}(\gamma_{\sigma_i}(\pi_{V_i}P)),
   \quad i=0,\ldots,m-1,
$$
for any box $P$ of type $V \to D$,
and call
$\gamma_i$ the contraction operator of $s_i$.
We define
$$\Gamma(P) = \gamma_0(P)
\cap \cdots \cap
\gamma_{m-1}(P),
$$
and call $\Gamma$ the contraction operator of the CSP.
\end{definition}

\begin{lemma}\label{lem:Gamma:properties}
$\Gamma$ is inflationary, monotonic, and correct.
\end{lemma}
\emph{Proof} 
Since, by Lemma \ref{lem:gamma:properties},
each $\gamma_{\sigma_i}$ is inflationary,
one has
\begin{eqnarray*}
  \Gamma(P) &=& \bigcap_{i=0}^{m-1} \gamma_i(P)
            = \; \Join_i \gamma_i(P) \\ 
&=& \Join_i \pi_V^{-1}(\gamma_{\sigma_i}(\pi_{V_i}P)) \\
&=& \pi_V^{-1} ( \Join_i (\gamma_{\sigma_i}(\pi_{V_i}P))) \\
&\sqsupseteq& \pi_V^{-1} ( \Join_i \pi_{V_i}P)) \\ 
&=& P
\end{eqnarray*}
Hence $\Gamma$ is \emph{inflationary.} 

$\Gamma$ is monotone, as a composition of monotone operators,
since both projection $\pi_{V_i}$ and cylindrification $\pi_V^{-1}$
are monotone operators.

Finally $\Gamma$ is correct.
Indeed, since by Lemma \ref{lem:gamma:properties}
each $\sigma_i$ is correct, i.e. satisfies
$\sigma_i \cap (\pi_{V_i} P \setminus \gamma_{\sigma_i} (\pi_{V_i} P) ) = \emptyset$,
one has, for any tuple $f$, that
$ f \in (P \setminus \Gamma(P)) \Leftrightarrow 
f \in P \mbox{ and } \exists i\ f \not \in \gamma_i(P)$
i.e., $f_{V_i} \not \in \sigma_i$ i.e., $ f \not \in \sigma$.
Hence
$ f \in (P \setminus \Gamma(P)) $
implies $ f \not \in \sigma$, 
thus 
$ \sigma \cap (P \setminus \Gamma(P)) = \emptyset $.
Therefore $\Gamma$ is correct.
                                     \hspace{\fill}$\Box$
\paragraph{}
A counter example to the idempotency of $\Gamma$
is given by the CSP example discussed earlier,
in Section \ref{sec:example}:
\begin{equation*}
\{\sq(x,y), \sq(y,z), \sm(y,z,u), \one(u)\}.
\end{equation*}
It is enough to take \emph{e.g.}
the approximation domain of (real)
boxes included in $ \set{x,u,z,u} \to \R$,
the corresponding $\Gamma$ operator operating on that domain, 
 together with the box $P$ informally described in equation
(\ref{eq:box2}), namely
$ P = \seT{f : \set{x,y,z,u} \to \R}{f(x) \in [\frac{1}{2},1],
f(y) \in [0,1],
f(z) \in [0,1],
f(u) \in [1,1].
}
$. 
The sequence $(\Gamma ^n(P))_{n\in\mathbb{N}}$
is strictly decreasing until it stabilizes at the smallest box, 
in the approximation domain, containing the tuple
$f : \set{x,y,z,u} \to \R$,
such that $f(x) = \surd (\frac{1}{2}(\surd 5 - 1)),
f(y) = \frac{1}{2} (\surd 5 -1 ),
f(z) = \frac{1}{4} (\surd 5 -1 )^2$,
and
$f(u) = 1$.

\subsection{Algorithms}

Algorithms for solving CSPs proceed 
by applying contraction operators.
Hence the algorithms only remove tuples from consideration
that are not part of the solution.
In the course of this process
absence of solutions of the CSP may be demonstrated,
but solutions are not, in general, constructed.

In the case of a discrete $D$ it may happen that
applying constraint contractors may result in a box
that contains a single tuple.
This tuple will then need to be substituted in the CSP
to check whether it is a solution.
However, in the type of CSP we are concerned with here
(reals with floating-point intervals as approximation domain),
finding a solution this way
is but a remote theoretical possibility
(the problem would have to have an exact solution
in terms of floating-point numbers, which, moreover,
upon substitution would miraculously avoid rounding errors).
Hence for numerical CSPs the best we can expect
is an algorithm that results in a small box.
This box can be small indeed:
in double-length IEEE-standard floating-point arithmetic
the box can have as projections intervals of relative width
around $10^{-17}$.
The result shows that, \emph{if} a solution exists,
it has to be in that box.

Among the algorithms that use contraction operators
to solve CSPs we distinguish two types of iteration
according to the order in which the operators are applied.
We distinguish \emph{rigid} order from and \emph{flexible} order.
The latter type leaves more choice
in the choice of the next operator to be applied.

Consider a CSP $\angTup{C,V,D,M}$ with contraction operators
$\gamma_0,\ldots,\gamma_{m-1}.$
The rigid-order algorithm applies the $m$ operators
in such an order that between two successive applications
of any particular operator all other operators are applied.
The rigid-order algorithm is susceptible to improvement.
In a typical CSP $m$ can be in the order of hundreds or thousands,
whereas each of the constraints typically has few arguments.
In numerical CSPs, for example, there are three or fewer.
Usually each constraint shares an argument with several others.
In such a situation
most of the contractor applications have no effect:
each application affects only few of many arguments
and it may well be that the next operator belongs to a constraint
that does not involve any of these few arguments,
so that its application has no effect. 

This suggests a chaotic algorithm,
one that avoids
such ineffectual choices of operator applications\footnote{
The term ``chaotic'' has been adopted by the constraint processing
literature via a detour from a numerical algorithm \cite{chzn69}.
}.
There is  considerable scope for such optimization,
as the only constraint on the sequence of operator applications
is that this sequence be \emph{fair} in the following sense.
\begin{definition}\label{def:fair}
Let $k \in (\nat \to A)$ be an infinite sequence
of which the elements are members of a finite set $A$.
$k$ is \emph{fair} iff each element of $A$ occurs
infinitely many times in $k$.
\end{definition}
Thus, in a fair sequence, it is possible,
but not necessary,
that between two occurrences of the same item
all other items have occurred.

A chaotic algorithm with $m$ operators applies the operators
in a fair sequence.
Such an algorithm can generate a fair sequence
while maintaining a record of the last index in the sequence
where a change was effected.
As soon as all the operators
have been applied without any resulting change,
then, by idempotence, the algorithm can be halted:
the rest of the infinitely long fair sequence
consists of operator applications that have no effect.
For details, see \cite{aptEssence}.

\subsection{Maximization property of the chaotic algorithm}

The chaotic algorithm solves the following problem:
\begin{equation}\label{eq:maxT}
   \left.
\begin{array}{ll}
   \mbox{{\bf maximize}} & B  \\
   \mbox{{\bf subject to}} & B \sbs \Gamma(B) 
\end{array}
   \right\}
\end{equation}
\noindent
where $B$ ranges over the boxes in the approximation domain, and
$\Gamma$ is the $\Gamma$ operator associated with the CSP.
The problem is stated in a format borrowed from
``mathematical programming'' in the sense that this includes,
for example, linear programming.
In the above format the total order among real numbers
has been replaced by the partial order
which is the Scott information order
described in Section~\ref{sec:notTerm}.
The generalization from the total order of mathematical programming
to programming with partial orders is due to Parker
who captures a wide variety of algorithms
in this framework \cite{prkr87}.

It is easily seen that chaotic iteration
solves the maximization problem if the sequence
generated by the algorithm converges
to the least fixpoint of  $\Gamma$.
Note that $\sbs$ is the information order,
where $B_0 \sbs B_1$
iff each of the projections of $B_1$ is a subset
of the corresponding projection of $B_0$.

\paragraph{Fixpoints}
We review some basic facts about fixpoints.
Let \mbox{$\langle D,\sbs,\bot \rangle$}
be a complete partially ordered set.
Completeness means here that every infinite ascending chain
$c_0 \sbs c_1 \sbs \ldots$ has a least upper bound
$\bigsqcup_{i=0}^\infty c_i$ that is an element of
the partially ordered set.

Let $\Gamma \in (D \to D)$ be monotonic and continuous.
Continuity of a function $f \in D \to D$ means
that for every infinite ascending chain
$c_0 \sbs c_1 \sbs \ldots$
we have
$f(\bigsqcup_{i=0}^\infty c_i) = \bigsqcup_{i=0}^\infty f(c_i)$.
In case of a finite $D$
such as the partially ordered set of floating-point intervals,
monotonicity implies continuity.
By the Knaster-Tarski theorem, $\Gamma$ has a least fixpoint
$\lfp(\Gamma) \in D$. 
This may be seen as follows.

By monotonicity of $\Gamma$,
$$
\bot \sbs \Gamma(\bot) \sbs \Gamma^2(\bot) \sbs \cdots
$$
By the completeness of the partially ordered set,
$\bigsqcup_{n=0}^\infty \Gamma^n(\bot) \in D$.
By the continuity of $\Gamma$,
$$
\Gamma(\bigsqcup_{n=0}^\infty \Gamma^n(\bot))
=
\bigsqcup_{n=0}^\infty \Gamma(\Gamma^n(\bot))
=
\bigsqcup_{n=0}^\infty \Gamma^n(\bot).
$$
Hence $\bigsqcup_{n=0}^\infty \Gamma^n(\bot)$
is a fixpoint of $\Gamma$.

%

We now turn to the Tarski fixpoint theorem.
Let $\Gamma \in (D \to D)$ be monotonic, but
assume now that partially ordered set
\mbox{$\langle D,\sbs,\bot \rangle$}
is a complete lattice, a richer structure. Completeness means here
that \emph{any} subset of $D$ has a least upper bound and a greatest
lower bound.  In particular $D$ possesses a largest element $\top$.
Then by the Tarski fixpoint theorem $\Gamma$ has a least fixpoint
$\lfp(\Gamma) \in D$. 
This may be seen as follows.

Consider the set
$S = \seT{a\in D}{\Gamma(a) \sbs a}$. $S$ is non-empty since it contains
top element $ \top \in D$. Let 
$l = \sqcap S$ be the greatest lower bound of $S$.
Then for any element $a \in S$, one has
$$ a \in S \Rightarrow l \sbs a \Rightarrow \Gamma(l) \sbs \Gamma(a) \sbs a $$
by monotonicity of $\Gamma$.
Hence $\Gamma(l)$ is lower bound for $S$, $ \Gamma(l) \sbs l = \sqcap S$.
Therefore $l \in S$.
One then has the chain of implications

$$ \Gamma(l)
\sbs l \Rightarrow \Gamma(\Gamma(l))
\sbs \Gamma(l) \Rightarrow \Gamma(l) \in S \Rightarrow l
\sbs \Gamma(l) \Rightarrow l = \Gamma(l).
$$
Hence $l$ is a fixpoint of $\Gamma$.
It is also the least fixpoint, since 
$S$ contains every fixpoint, and $l = \sqcap S$.
Therefore $l= \sqcap S = \lfp(\Gamma)$ is the least fixpoint of $\Gamma$.

\paragraph{Application of fixpoint theory
to the chaotic algorithm}

\begin{theorem}\label{thm:probeAppr}
Given a CSP $\angTup{C,V,D,M}$
with contraction operator $\Gamma$ and solution set $\sigma$.
For any box $P$ of type $V \to D$ we have
$$
(\sigma \cap P)
\subseteq
\Box(\sigma \cap P)
\subseteq
\Gamma^n(P)
$$
for all $n = 0,1,2,\ldots$
\end{theorem}
\emph{Proof}
The first inclusion follows from the definition of the $\Box$
operator.
We consider the case where there are $m = 2$ constraints,
which easily extends to arbitrary greater values of $m$.
We write $\sigma_i = M(s_i)$
and $V_i$ for the set of variables in $s_i$,
for $i = 0,1$.
We first consider the case $n = 1$.
\begin{eqnarray*}
\Box(\sigma \cap P) &=& \\
\Box\seT{a \in (V \to D)}
  {a_{V_0} \in \sigma_0 \wedge
    a_{V_1} \in \sigma_1 \wedge a \in P}
                                       &=& \\
\Box\set{a \in (V \to D) :
  a_{V_0} \in \sigma_0 \wedge
    a_{V_1} \in \sigma_1 \wedge
      a_{V_0} \in \pi_{V_0}P \wedge
        a_{V_1} \in \pi_{V_1}P        } &=& \\
\Box\set{a \in (V \to D) :  
  a_{V_0} \in (\sigma_0 \cap \pi_{V_0}P) \wedge
   a_{V_1} \in (\sigma_1 \cap \pi_{V_1}P) } &=& \\
\Box(\pi_V^{-1} (\sigma_0 \cap \pi_{V_0}P) \cap 
  \pi_V^{-1} (\sigma_1 \cap \pi_{V_1}P))              &\subseteq&\\
\Box(\pi_V^{-1} \Box(\sigma_0 \cap \pi_{V_0}P) \cap
  \pi_V^{-1} \Box(\sigma_1 \cap \pi_{V_1}P))              &=&\\
\pi_V^{-1} \Box(\sigma_0 \cap \pi_{V_0}P) \cap
  \pi_V^{-1} \Box(\sigma_1 \cap \pi_{V_1}P)          &=&\\
  \gamma_0(P) \cap
    \gamma_1(P)              &=&\\
    \Gamma(P).              &&\\
\end{eqnarray*}
We have shown that
$ \Box(\sigma \cap P) \subseteq \Gamma(P). $
We also have $ \Box(\sigma \cap P) \subseteq \Gamma^2(P).$
This is because of the correctness of $\Gamma$:
it does not remove any solution tuples from its argument.
Hence we have $ \Box(\sigma \cap P) \subseteq \Gamma^n(P)$
for any $n \geq 0$.

\hspace{\fill}$\Box$

By Definition~\ref{def:bigGamma},
$\Gamma$ is the intersection of contraction operators,
one for each constraint,
each of which can be efficiently computed.
The results of these operators are exact
in the sense that the results are by definition approximations
and are therefore exactly representable.
Thus Theorem~\ref{thm:probeAppr} can serve as the basis for an algorithm
for approximating the set of solutions in $P$.

In terms of the information order $\sbs$ Theorem~\ref{thm:probeAppr}
states that
$\Gamma^n(P) \sbs \Box(\sigma \cap P) \sbs (\sigma \cap P)$. 
\begin{theorem}
$\Gamma$ is monotonic on the partially ordered
set of subboxes of $P$ ordered by information order.
\end{theorem}
\emph{Proof}
  Each contraction operator
$\gamma_i :
  P \mapsto \pi_V^{-1}(\gamma_{\sigma_i}(\pi_{V_i}P))
$
is monotone, and the join of two monotone operators is
monotone.

\hspace{\fill}$\Box$

\paragraph{}
Observe that the set of boxes contained in $P$ defines an approximation 
structure for $P$.
$\Gamma$ is monotonic.
The partially ordered set of subboxes of
$P$ is ordered by information order
and is a complete lattice with least element $P$. 
Accordingly,
$\Gamma$, restricted to the approximation structure,
has a least fixpoint $\lfp(\Gamma)$,
by the Tarski fixpoint theorem.
Summarizing, we have
$\Gamma^n(P) \sbs
 \lfp(\Gamma) \sbs
 \Box(\sigma \cap P) \sbs
 (\sigma \cap P)
$
for all $n$.

If the box operator $\Box$ is continuous over the approximation domain
defined over $D$, then $\Gamma$ is also
continuous by compositionality of continuous functions, and
by the Knaster-Tarski theorem 
$\bigsqcup_{i=0}^\infty \Gamma^i(P) $
is the least fixpoint of $\Gamma$ contained in $P$.

In particular, if $D$ is the set $F$
of finite double-length IEEE-standard floating-point numbers,
and the approximation domain is given by the set of $F$-intervals,
then domain $D$ is finite, hence both operators $\Box$ and $\Gamma$
are continuous.
The subboxes of $P$ form a complete partially ordered set
trivially because the finiteness of the set of floating-point
numbers.
Therefore
$\bigsqcup_{i=0}^\infty \Gamma^i(P) 
= \bigsqcup_{i=0}^n \Gamma^i(P) 
$,
for some finite $n$, 
is the least fixpoint of $\Gamma$, restricted to $P$.



\begin{theorem}
Let a CSP
$\angTup{\set{s_0,\ldots,s_{m-1}},V,D,M}$, with
contraction operator $\Gamma$, and contraction operators 
$\gamma_i$ for each individual constraint $s_i$ be given.
If the approximation structure over $D$ is such that
the box operator $\Box$ is continuous, then,
for every box $P$,
every fair iteration of continuous operators $\gamma_i$ starting with
$P$ converges towards the least fixpoint 
$ \sqcup_{j=0}^\infty \Gamma ^j(P) $
of $\Gamma$, restricted to $P$.
\end{theorem}

\emph{Proof}
Let 
$k_0, k_1, k_2, \ldots$
be a fair iteration, where for each $n$,
$ k_n \in \set{0,\ldots,m-1}
$
is the index of the constraint $s \in \set{s_0,\ldots,s_{m-1}} $
selected at the $n$th iteration step.
The corresponding iteration starting from some box $P$ is given by
the sequence of boxes
\begin{eqnarray*}
  P_0 &=& P \\
  P_n &=& \gamma_{k_n} (P_{n-1}),  \qquad n > 0
\end{eqnarray*}

We first show that
\begin{equation}\label{eq:bound1}
\forall j\ \exists q\ \Gamma ^j (P) \sqsubseteq P_q 
\end{equation}
Indeed,
$k$ is a fair sequence, and since all operators $\gamma_i$ are
inflationary and monotone, for each $j$,
one can choose $q$ such that the initial iteration subsequence
$k_0, \ldots, k_{q-1}$
contains, for each constraint $s_l$ in $C$, 
at least $j$ occurrences of index $l$ of $s_l$ in 
$\set{0,\ldots,m-1}$; these occurrences
correspond to at least 
$j$ applications of the contraction operator $\gamma_l$.

Next, we observe that
\begin{equation}\label{eq:bound2}
\forall q\ P_q \sqsubseteq \Gamma ^q (P),
\end{equation}
which follows by induction on $q$.

Whence 
$\sqcup_{j=0}^\infty \Gamma ^j(P) \sqsubseteq \sqcup_{j=0}^\infty P_j$
by (\ref{eq:bound1}), and
$\sqcup_{j=0}^\infty P_j \sqsubseteq \sqcup_{j=0}^\infty \Gamma ^j(P) $
by (\ref{eq:bound2}).
The two limits are equal.
\hspace{\fill}$\Box$

\section{Further work}
\label{sec:furthWrk}

Concurrent constraint programming (CCP)
(\cite{srrnpn91} and further references there)
is a model of concurrent programming.
This model is based on an abstraction of a computer
store that is more abstract than the one
used in conventional programming languages.
Usually the store is modeled as a vector of storable values
(numbers, characters) indexed by the variables accessible to
the program.
Thus to every variable there corresponds a single value.
The conventional read operation on a variable yields this value.
The conventional write operation on a variable changes this value.

In CCP it is not assumed that the value of a variable
is precisely known:
the store is a \emph{constraint} on the values of variables.
The conventional read operation
is replaced by {\tt ask}, an operation in the form of a logic formula
that succeeds if and only if it is logically entailed by the store.
The conventional write operation
is replaced by {\tt tell}, an operation in the form of a logic formula
$T$
that has the effect of replacing the store $S$ by a logical
equivalent of $S \wedge T$, provided that this is consistent. 

The generalization of the conventional store to CCP
requires that the store becomes a logical theory $S$
that is \emph{satisfaction-complete}
in the sense that for every formula $C$
admissible as {\tt ask} or {\tt tell}
it is the case that either
$ S \models \exists C$ or
$ S \models \neg \exists C$
where $\exists$ denotes existential closure.
See \cite{clrk91} and further references there.

CCP seems to have a great deal of unexploited potential.
Its motivation and terminology is in the area of concurrent
programming, with the aim of generalizing the many
different approaches
(Hewitt's Actors, Hoare's CSP, Milner's CCS, various flavours
of concurrent logic programming).
CCP is linked to constraint solving
by its formulation in terms of predicate logic.
Thus CCP promises to be a framework for constraint solving
with parallelism built in, a promising feature given
the massive amount of computation that is typical of
constraint problems.

To realize this promise it is necessary to generalize
CCP beyond the restriction of the store
as a satisfaction-complete theory.
For example, in the case of interval constraints,
where the domain is the reals,
the theory of the store is not satisfaction-complete.
Consequently, the result of a converging iteration with
interval constraints means that \emph{if} a solution
exists, then it has to be in the remaining intervals.
Often one knows from other sources that a solution exists
(e.g. that the CSP arises from a polynomial of odd degree being
equated to zero) and the remaining intervals are close to
the resolution of the floating-point system.
In such a situation the weakness of the conclusion
does not stand in the way
of it being of great practical value.
We have not explored whether the valuable features
of CCP can be preserved when the store
is not a necessarily a satisfaction-complete theory.

\section{Concluding remarks}
\label{sec:Conclusion}

We see the contributions of this paper as the following.

Although in the usual definition of CSP the constraints
look like atomic formulas of predicate logic,
the semantics of a CSP is given independently.
We use the standard semantics of first-order predicate logic
to define the solution set of a CSP
and we define approximation systems as a set-theoretic device
to interface our framework for CSPs
with the well-known chaotic iteration algorithm.

Parker's observation \cite{prkr87} was that the operations research
paradigm of maximizing a real-valued objective function
under constraints can be generalized to maximization
in partially ordered spaces.
Scott's contribution \cite{scott72} was
that computation can be viewed as information gain.
We combine these insights, so that many of Parker's examples
can be seen as iterations in which information is monotonically
gained.

Among these examples we concentrate on solving systems
where the constraints are nonlinear
equations or inequalities over the reals.
Constraint processing by
domain reduction can be viewed as the use of the computer
for monotonic gain of information.
This is more than a theoretical point of view.
What is lacking in the current practice of computing
is a quantitative treatment of the \emph{work} done by
the cpu per, say, gigacycle. 
The domain reduction method can be used to compare how many
gigacycles were required to obtain the most recent domain reduction,
expressed, say, as ratio of the cardinalities, or volumes,
of the box before and after this reduction.
One may conclude that a reduction of $x$ percent is not worth
the $y$ gigacycles it cost, that further diminishing returns
for computational effort are to be expected, and that therefore
it is time to terminate the iteration.

\section{Acknowledgments}

This research was supported by our universities,
by INRIA Rocquencourt, France,
and by the Natural Science
and Engineering Research Council of Canada.

\bibliographystyle{abbrvnat}


\end{document}